\documentclass{article}

\usepackage{microtype}
\usepackage{graphicx}
\usepackage{subfigure}
\usepackage{booktabs} 

\usepackage{hyperref}

\usepackage[accepted]{icml2024}

\usepackage{amsmath}
\usepackage{amssymb}
\usepackage{mathtools}
\usepackage{amsthm}

\usepackage[capitalize,noabbrev]{cleveref}

\usepackage{booktabs}
\usepackage{enumitem} 
\usepackage{multirow}
\usepackage{multicol}

\usepackage{adjustbox}
\usepackage{subcaption}

\usepackage{dcolumn}

\usepackage{float}

\usepackage{amsmath}

\DeclareMathOperator*{\argmin}{arg\,min}

\usepackage{xspace}

\makeatletter
\DeclareRobustCommand\onedot{\futurelet\@let@token\@onedot}
\def\@onedot{\ifx\@let@token.\else.\null\fi\xspace}

\newcommand{\myparagraph}[1]{\vspace{4pt}\noindent\textbf{#1.}}
 
\def\ie{\emph{i.e}\onedot}

\makeatother

\theoremstyle{plain}

\theoremstyle{definition}

\theoremstyle{remark}

\usepackage[textsize=tiny]{todonotes}

\icmltitlerunning{Scale-Free Image Keypoints Using Differentiable Persistent Homology}

\begin{document}

\twocolumn[
\icmltitle{Scale-Free Image Keypoints Using Differentiable Persistent Homology}

\icmlsetsymbol{equal}{*}

\begin{icmlauthorlist}
\icmlauthor{Giovanni Barbarani}{disma}
\icmlauthor{Francesco Vaccarino}{disma}
\icmlauthor{Gabriele Trivigno}{dauin}
\icmlauthor{Marco Guerra}{cnrs}
\icmlauthor{Gabriele Berton}{dauin}
\icmlauthor{Carlo Masone}{dauin}
\end{icmlauthorlist}

\icmlaffiliation{disma}{Department of Mathematical Sciences
"Giuseppe Luigi Lagrange", Politecnico di Torino, Italy}
\icmlaffiliation{dauin}{Department of Control and Computer Engineering, Politecnico di Torino, Italy}
\icmlaffiliation{cnrs}{Institut Fourier, Université Grenoble Alpes, France}

\icmlcorrespondingauthor{Giovanni Barbarani}{giovanni.barbarani@gmail.com}

\icmlkeywords{Machine Learning, ICML}

\vskip 0.3in
]

\printAffiliationsAndNotice{}

\begin{abstract}

In computer vision, keypoint detection is a fundamental task, with applications spanning from robotics to image retrieval; however, existing learning-based methods suffer from scale dependency, and lack flexibility. This paper introduces a novel approach that leverages Morse theory and persistent homology, powerful tools rooted in algebraic topology. We propose a novel loss function based on the recent introduction of a notion of subgradient in persistent homology, paving the way toward topological learning. Our detector, MorseDet, is the first topology-based learning model for feature detection, which achieves competitive performance in keypoint repeatability and introduces a principled and theoretically robust approach to the problem.

\end{abstract}

\section{Introduction}
\label{introduction}

The ability to extract salient points (\emph{keypoints}) and associated features from an image is a cornerstone of computer vision, as it underpins several applications such as visual localization \cite{sarlin2019coarse,sattler2018benchmarking,Toft2020TPAMI}, SLAM \cite{artal2015orb,durrant2006simultaneous,bailey2006simultaneous2}, Structure-from-Motion and 3D reconstruction \cite{schoenberger2016sfm,heinly2015CVPR,schoenberger2016mvs}, as well as retrieval and place recognition \cite{barbarani2023local,noh2017delf}. 
Traditional pipelines relied on handcrafted filters that were engineered to detect salient points such as corners \cite{harris1988corners}, blobs \cite{tuytelaars2000wide,lowe2004distinctive,mikolajczyk2004affine} or edges \cite{bhardwaj2012edge}. These keypoints would then be associated with a feature vector obtained typically from local derivatives of the image \cite{calonder2010brief,bay2006surf,lowe2004distinctive}. 

Ideally, a good feature detector should provide keypoints with the following desirable properties: high
repeatability (\ie, consistent across image pairs) and scale-invariance, while being robust to noise and distortion \cite{ghahremani2020ffd,revaud2019r2d2,lowe2004distinctive}. Scale-Space theory \cite{lindeberg1994scale} provides a formulation of the concept of keypoint that guarantees the properties mentioned above \cite{lindeberg1994scale,lowe2004distinctive,ghahremani2020ffd}, and it operates by building a scale-space feature pyramid from the image, in which keypoints are detected as local extrema. Many classical handcrafted detectors exploit this theoretical framework \cite{mikolajczyk2004affine,bay2006surf}, the most popular of which is SIFT \cite{lowe2004distinctive}. 

Recently, several learning-based detectors have been introduced, which, in the spirit of deep learning, propose to forego the formal definition of keypoints and rely on a data-driven approach to teach a neural network how to select salient points \cite{yi2016lift,savinov2017quad,tian2017l2net,revaud2019r2d2}.
Despite their learning-based approach, these methods are still bound to several design choices: for reliability, keypoints are often defined as locations that are easily matched (DISK) or that are more discriminative (R2D2); and for repeatability, they are usually defined as local maxima, in a local patch of arbitrary size. 

This formulation presents some inherent flaws due to the coarse heuristic used to model the maxima (\ie, patch-wise) being inflexible.
More specifically it
(i) does not guarantee that chosen points are critical points of the feature maps, 
(ii) leads to keypoints whose density is hard-coded in a hyperparameter, which
(iii) makes the keypoints inherently scale-dependent.
Whereas the issue of scale dependency is often approached through multi-scale inference (\ie, processing the same image multiple times at different resolutions).
We note that a mathematical framework that is able to model and thus locate local maxima with inherent guarantees of scale independence is currently missing in the literature. 

To this end, we propose a novel and differentiable formulation of keypoints based on Morse theory \cite{milnor1963morse} and persistent homology \cite{edelsbrunner2002topological, zomorodian2004computing} from algebraic topology. This formulation leverages the connection between local maxima and differentiable topological invariants \cite{carriere2021optimizing, leygonie2021framework} offering a more robust and elegant solution, thus enabling a genuinely unsupervised framework for keypoint detection, thus
without requiring hardcoded hyperparameters that determine the density or frequency of keypoints.
We believe this work can lay a foundation for future research to expand our proposed framework for integrating topology in computer vision. Our implementation and trained models have been publicly released\footnote{\url{https://github.com/gbarbarani/MorseDet}}.

To summarize, our contributions are the following:
\begin{itemize}[noitemsep,topsep=1pt]
    \item We show a connection between topological data analysis (TDA) and keypoint detection. Our method, MorseDet, is the first learnable method based entirely on a TDA framework;  
    \item The first loss based on persistent homology for unsupervised learning, capable of modeling a set of features arbitrary in cardinality and shape;
    \item We demonstrate that our detector, thanks to its strong theoretical foundation, shows promising performance in terms of keypoints repeatability. 
\end{itemize}

\begin{figure*}[!t]
    \centering
    \adjincludegraphics[width=\linewidth, height=4.5cm, trim={{.01\width} {.2\height} {.1\width} {.2\height}}, clip]{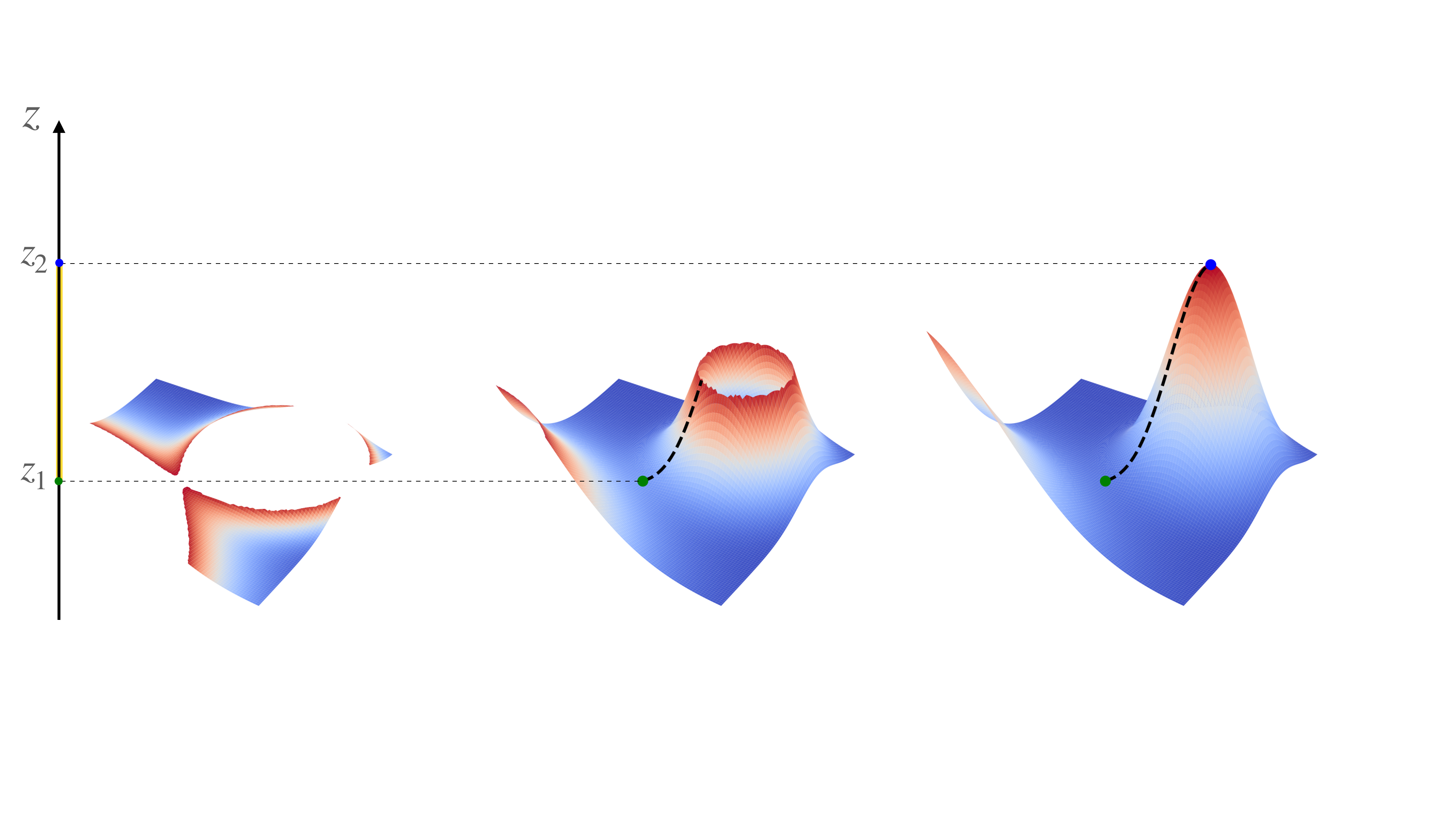}
    \caption{\label{fig:MorseFiltration} The evolution of the sub-level sets of a surface filtered by height, \ie value on the $z$ axis. As the height crosses $z_1$, a new loop is born in correspondence with a saddle (green point), then the loop changes smoothly until $z$ hits  $z_2$, the value of a corresponding maximum (blue point), and the loop disappears. $z_1$ and $z_2$ are, respectively, the topological feature's birth time and death time.}
\end{figure*}

\section{Related Work}

\paragraph{Feature detection.} 

The task of local feature detection and description has witnessed a growing interest in the past years due to its central role in several vision-related applications~\cite{csurka2018handcrafted,salahat2017recent}. Traditionally, the detection step was performed independently via handcrafted methods based on local derivatives such as \cite{lowe2004distinctive,harris1988corners}. Early attempts to train learnable detectors also adopted this decoupled approach \cite{mishkin2018repeatability,barroso2019key}.
The common denominator of these early works is that they all rely on a supervised notion of keypoint, which is problematic because it is limited by the prior knowledge that the researchers have on what should be considered a keypoint, either via handcrafted detectors or via learnable filters that mimic the handcrafted ones. SuperPoint \cite{detone2018superpoint} was the first to implement a semi-supervised approach. Starting from synthetic shapes with user-defined keypoints, it uses homographic adaptation to enhance the equivariance of the detected keypoints to such transformation. This approach is still limited because keypoints are still arbitrarily defined as intersections of known structures.

D2-Net \cite{Dusmanu2019CVPR} and R2D2 \cite{revaud2019r2d2} proposed a joint detection and description pipeline, arguing that keypoint repeatability cannot be achieved via supervised training, which would result in mimicking a previously available detector. They are directly inspired by Scale-Space theory \cite{lindeberg1994scale} and model keypoints as local maxima of their output maps.  
More recent self-supervised approaches, namely SiLK \cite{Gleize_2023_ICCV} and DISK \cite{tyszkiewicz2020disk}, avoid the problem of defining keypoints, and promote the detection of points that are easy to match and are based on probabilistic loss formulations. 
ALIKE \cite{zhao2023alike} proposes a patch-wise softmax relaxation of keypoints at training time. 
ALIKED \cite{zhao2023aliked} further improves this approach by relying on deformable convolutional filters capable of adapting to the keypoints support.

These recent works demonstrated the effectiveness of a learnable approach based on detecting local maxima from the response map of a feature extractor. Nevertheless, they lack an analytical tool to locate these local extrema reliably and rely typically on a softmax-based approximation inside local patches of predefined size, thus bounding the detection frequency to this hyperparameter and fails to achieve scale invariance. We propose a formulation based on the persistent homology theory that fills this gap in the literature. 

\paragraph{Topological data analysis.} Topological data analysis (TDA) encompasses several data analysis techniques that employ algebraic topology. These include persistent homology \cite{edelsbrunner2002topological}, which is often utilized to provide a multi-scale description of point clouds, and discrete Morse theory \cite{delgado2014skeletonization,robins2011theory}, used extensively for processing 2D and 3D images. Persistence-based statistics have proven useful in various domains, such as clustering \cite{chazal2013persistence}, robust pose estimation \cite{dey2010persistent}, and as input features for neural networks \cite{hofer2017deep,giansiracusa2019persistent}. Although in a non-learnable context, the Morse theory and TDA have been applied to image matching \cite{MATAS2004761, 6467250, 6940260}. From a theoretical standpoint, the differentiability of persistence-based functions has been extensively explored \cite{leygonie2021framework,carriere2021optimizing}. Recently, differentiable topological objectives have been integrated into deep learning, serving as regularization tools for shaping the topology of decision boundaries in classification tasks \cite{chen2019topological} and as priors for the latent space in autoencoders \cite{moor2020topological}. In supervised learning contexts, persistent homology, along with discrete Morse theory, has been applied to image segmentation tasks to enhance the loss function \cite{hu2023learning,hu2021topology,clough2020topological}, often complemented by the binary cross-entropy loss, and to quantify uncertainty \cite{gupta2024topology}.

Our work represents a first attempt at introducing completely unsupervised objectives based on persistent homology within the context of deep learning for data-intensive applications.

\section{Background}

\begin{figure*}[t]
    \centering
    \includegraphics[width=\textwidth, height=5cm, keepaspectratio]{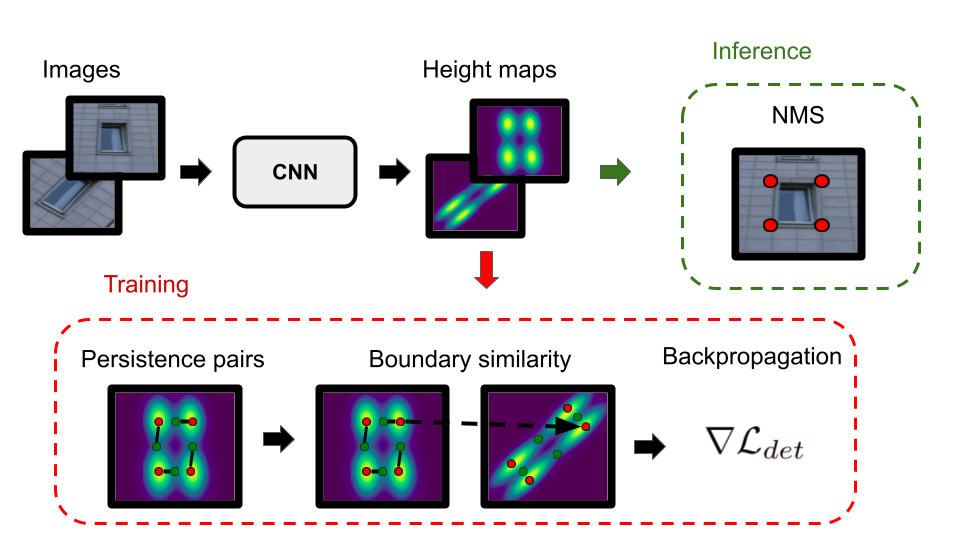}
    \caption{\label{fig:pipeline}   \textbf{Pipeline overview}: a convolutional neural network (CNN) is employed to generate height maps from input images. During inference, keypoints are efficiently detected as local maxima of these maps, utilizing non-maximum suppression. For training, our detector loss computation involves the application of discrete Morse theory algorithms to compute persistence pairs, and then the maps at the corresponding positions are compared through the boundary similarity. The resulting gradients are subsequently backpropagated.}
\end{figure*}

The relationship between critical points of a function (extrema and saddle points) and the evolution of a topology can be intuitively explained using the following analogy: picture the graph of a 2D scalar function as a landscape. When we flood this landscape, we witness a series of transformations: lakes emerge from the lowest valley regions; lakes surround mountains, leaving only their peaks above water, and, ultimately, the lakes blend when they submerge the peaks.  

Morse theory \cite{milnor1963morse} is the mathematical framework that precisely captures the relationship between critical points and changes in topology, while persistent homology \cite{edelsbrunner2002topological} is the algebraic topology tool that defines the computation of the filtration process described above. Finally, discrete Morse theory \cite{robins2011theory} is the tool that is commonly used to implement these ideas in the context of digital 2D and 3D images.
We will briefly review these concepts in the context of our scope.

\subsection{Morse Theory}
\label{sec:morse_theory}
A smooth scalar function $h$ defined on a smooth manifold is a Morse function if it has only non-degenerate critical points, \,  i.e., having non-zero Hessian determinants only. This condition is not restrictive: indeed, up to an infinitesimal perturbation, every differentiable function on a compact is Morse. 
Given a 2D compact surface $\mathcal{X}$ and the choice of a Morse function $h$, we can study the evolution of the sub-level sets $\mathcal{X}_t = \{x\in \mathcal{X}: h(x) \le t\}$ for an increasing $t$. These sets can be considered the union of the bottom of the lakes obtained by pouring water onto our landscape up to level $t$. 
When $t$ reaches the value corresponding to a minimum of $h$, the sub-level changes by adding a new point: a new connected component (lake) is born. 
When $t$ reaches a saddle point $s=(p,t)$ with $t=h(p)$, two things could happen: (i) the saddle $s$ merges two lakes into one, or (ii) the saddle $s$ creates a single span bridge over a lake, thus producing a new closed path (loop) in the component.
Therefore, a saddle either reduces connected components or creates a loop.
Finally, when $t$ reaches a maximum value, it corresponds to completely submerging the terrain and its closed paths, and this can be seen as filling the hole surrounded by a closed path. 

\subsection{Persistent Homology}

The homology modules $H_0$ and $H_1$ of $\mathcal{X}$ are vector spaces, whose dimensions count the number of connected components and the number of loops of $\mathcal{X}$, respectively. The sub-level filtration $\mathcal{X}_{t_1} \subset \mathcal{X}_{t_2} \subset ... \subset \mathcal{X}_{t_n}$ associated to a function $h$ in an obvious way, induces a sequence of morphisms 
\begin{equation*}
H_i(\mathcal{X}_{t_1}) \to H_i(\mathcal{X}_{t_2}) \to ... \to H_i(\mathcal{X}_{t_1})
\end{equation*}

The persistent homology module $\mathbb{H}_i(\mathcal{X},\mathcal{F}):=\bigoplus_t H_i(\mathcal{X}_{t})$ of $\mathcal{X}$ keeps track via the morphisms above of the evolution along the sub-level filtration of the topological features associated to the critical points of the function $h$. Each generator of $e\in\mathbb{H}_i$  (as a graded module over the ring of polynomials, see app. \ref{sec:algtop}) is associated with a pair of values $(b(e),d(e))$, $b(e)<d(e)$, the birth time and the death time of $e$, representing the life span of a topological feature. 
From sec. \ref{sec:morse_theory} it should be apparent that, if $e \in \mathbb{H}_0$, $b(e)$ corresponds to a minimum of $h$ and $d(e)$ must be a saddle. On the other hand, if $e \in \mathbb{H}_1$, then $b(e)$ corresponds to a saddle point and $d(e)$ to a maximum; an example of this case is depicted in fig. \ref{fig:MorseFiltration}.

\subsection{Discrete Morse Theory}

To study the topology of the local extreme values of a digital image $I$, the cubical complex $\mathcal{K}$ and the filtration associated to $I$ is used (see app. \ref{sec:cubical}). Discrete Morse theory is an efficient framework to compute the birth and death times of generators of the associated modules $\mathbb{H}_0$ and $\mathbb{H}_1$ through simulated differentiation. Also in this case the critical times are, up to a perturbation, unequivocally associated with the local extreme values of $I$.

\section{Methodology}

\myparagraph{Overview}
We propose a novel keypoint detector, which we call \textbf{MorseDet}, whose goal is to detect a set of sparse keypoints that should be repeatable across different transformations of an image as well as scale-invariant. The core novelty of our method is to utilize elements of topological data analysis, namely the persistent homology framework, to model keypoints as local maxima of the feature maps in a CNN in a differentiable way.
Our approach differentiates from previous works \cite{Dusmanu2019CVPR,revaud2019r2d2,zhao2023aliked} where local maxima are computed heuristically within a fixed \( N \times N \) patch, imposing a strong prior on the frequency of the detected keypoints, which ultimately makes the detection scale-dependent.
In MorseDet, we have re-envisioned these concepts with a scale-independent topological characterization of local maxima. We show that thanks to the theoretical guarantees that underpin our method, we outperform previous sparse detectors in terms of repeatability and scale invariance. Given the novelty of our topology-based approach, in this work, we focus only on the detection step, leaving for future works the integration of a topological descriptor loss.

\myparagraph{Problem setting}
We aim to learn a feature extractor $F_{\theta}$ that, given an input image $I \in \mathbb{R}^{H \times W \times C}$, outputs a set of discrete pixel locations $\{k_i\} \in \mathbb{R}^2$. 
Our backbone of choice is a simple fully convolutional network; in particular, we adopt the L2Net \cite{tian2017l2net} with the modification proposed in \cite{revaud2019r2d2} that employs smaller kernels in the last layers, in order to reduce its computational cost. 
Since we want to obtain a scalar map of the image, in order to exploit Morse theory, 
we modify the last layer of the backbone used in R2D2 \cite{revaud2019r2d2} to output a single channel.
Thus, forwarding an image through the adopted network, we obtain a response map $F_{\theta}(I) = \frak{H} \in \mathbb{R}^{H \times W}$. Essentially, the last layer distills the feature volume into a single-channel unified spatial representation, which we call \textit{height map}, in an analogy with the terminology used in Morse theory.

\subsection{Topological Detector Loss}

During the training process, we model keypoints bijectively with the local maxima of the feature map. The main novelty of our method is implementing a framework that guarantees to find all the extrema. We \emph{refer} to a local maximum via the associated topological feature, \,  i.e., the loop that spawns around the critical point and that gets closed at its peak. Formally, we consider the set $\mathcal{G}(\frak{H})$ of generators of $\mathbb{H}_1$, the $1$-dimensional persistent homology module of the cubical complex, with the filtration given by the height map $\frak{H}$ (see app. \ref{sec:cubical}). The set $\mathcal{G}(\frak{H})$ is in bijection with the set of local maxima of the height map, and its elements should be understood as the loops described above. Every element $ e \in  \mathcal{G}(\frak{H})$ can be associated to the coordinates of a (creator) saddle $s(e) \in \mathbb{R}^2$ and a (destructor) maximum $m(e) \in \mathbb{R}^2$. The birth time of $e$ is the value attained by $\frak{H}$ at its creator saddle, \ie, $b(e)=\frak{H}[s(e)]$, in the same way, the death time of $e$ is the value of a local maximum $d(e)=\frak{H}[m(e)]$. We make use of the \textbf{persistence} of $e$, a common way to measure the magnitude of a topological feature that is defined as 

\begin{equation}
  Pers(e)=d(e)-b(e)
\end{equation}

Notice that this quantity does not depend on the shape or extension of the region filled by the maximum, \ie, the scale, but only on how prominent the peak is. Given a map $U$ of (possibly sparse) correspondences between two height maps, $\frak{H}_1$ and $\frak{H}_2$, for convenience, we define the error matrix as

\begin{equation}
  E[i,j]=\frak{H}_1[i, j] - \frak{H}_2[U[i, j]]
\end{equation}

if $U$ is defined on $(i,j)$, otherwise $E[i,j]=0$. At this point, we introduce a new term that takes into account how the maps $\frak{H}_1$ and $\frak{H}_2$ differ at the topologically relevant (correspondent through $U$) positions, namely the \textbf{boundary similarity}:

\begin{equation}
  Sim(e) = E[s(e)]^2 + E[m(e)]^2
\end{equation}

Given a positive constant $\alpha$, our \textbf{detector loss} is finally defined as

\begin{equation}
\mathcal{L}_{det}(\frak{H}_1, \frak{H}_2)=-\sum_{e \in  \mathcal{G}(\frak{H}_1)} Pers(e) \left[ Pers(e) - \alpha Sim(e) \right]
\end{equation}

Let us now illustrate this object: the persistence can be thought of as a \emph{peaky term}, as indeed maximizing the persistence can be achieved by increasing the prominence of the peaks or their number. The boundary similarity can be considered a \emph{penalty term}, as it penalizes those topological features that are not reproducible across similar images. The hyperparameter $\alpha$ controls the trade-off of this regularization. 
Notice that, in contrast to patch-wise methods, we address the genuine set of local maxima found in the height map, which shows a high degree of variability and freedom. In this case, the persistence also takes on the role of a \emph{weight term} that multiplies the total contribution of the feature to the loss, and this prevents the optimization process from getting overwhelmed by thousands of noisy, low-intensity peaks, instead focusing on the refinement of the promising features. Formally, the loss is a quadratic function in the persistence and boundary similarity variables, where the interaction term introduces a convenient relation between their gradients.

The differentiability of our loss function follows from well-established results \cite{leygonie2021framework,carriere2021optimizing}, see app. \ref{sec:diff} for some details. A straightforward implementation within common deep learning automatic differentiation frameworks provides a valid gradient.

\begin{table*}[!htbp]
\begin{center}
\begin{adjustbox}{width=0.8\linewidth}
\begin{tabular}{lcccccccccccc}
\toprule
\multirow{2}{*}{{\begin{tabular}[c]{@{}c@{}}Method\end{tabular}}} & & \multicolumn{5}{c}{Illumination}  & & \multicolumn{5}{c}{Viewpoint}\\
\cline{3-7} \cline{9-13} 
&  & 250 & 500 & 1000 & 2000 & 4000 & & 250 & 500 & 1000 & 2000 & 4000  \\
\hline
D2-Net & & 21.1 & 22.0 & 23.6 & 26.4 & 28.7 & &  12.1 & 13.6 & 19.5 & 18.6 & 22.1\\
 R2D2   & & 27.3 & 28.6 & 29.8 & 30.5 & 30.7 &  & 24.3 & 25.5 & 26.5 & 27.6 & 28.3 \\
 SIFT       & & 34.9 & 37.2 & 38.8 & 40.4 & 41.2 & & \underline{37.8} & \underline{38.9} & 39.9 & 40.7 & 40.4\\

 SuperPoint & & \underline{42.4}  & \textbf{47.7} & \underline{49.8} & 49.5 & 49.4 & & 27.5 & 36.0 & \underline{43.6} & \textbf{46.8} & 46.4 \\
 DISK       & & 42.2  & 45.9 & \underline{49.8} & \textbf{54.2} & \textbf{57.4} & & 30.6 & 35.0 & 39.3 & 44.0 & \textbf{47.6} \\
ALIKED      & & 14.8  & 24.4 & 37.3 & 47.0 & 51.9 & &  6.5 & 10.6 & 18.1 & 29.7 & 43.1 \\

 \textbf{MorseDet (ours)}  & & \textbf{44.3} & \underline{47.3} & \textbf{50.3} & \underline{53.4} & \underline{55.2} & & \textbf{40.6} &  \textbf{42.8} & \textbf{44.6} & \underline{46.1} & \underline{47.2} \\
\bottomrule
\end{tabular}
\end{adjustbox}
\caption{Repeatability for illumination and viewpoint splits of HPatches, computed using various values for the maximum number of keypoints allowed. The \textbf{best} and \underline{second-best} results are indicated in each column.}
\label{tab:hprepeat}
\end{center}
\end{table*}

\subsection{Training and Inference}

For training our detector, we adopt WASF, the dataset released in \cite{revaud2019r2d2} to train R2D2, which provides homographic correspondences between pairs of images. 
A training instance is composed of two images $I_1, I_2$ and a ground-truth correspondences map between them $U \in \mathbb{R}^{H\times W \times 2}$, more explicitly $U[i, j] = (i', j')$ if and only if the pixel $(i', j')$ of the second image corresponds to the pixel $(i, j)$ in the first image; notice that $U$ is defined only on covisible regions.

The final training loss given a pair of images is the following:
\begin{equation}
\mathcal{L}(I_1, I_2) =  \mathcal{L}_{det}(\frak{H}_1, \frak{H}_2)
\end{equation}
Where $\mathcal{L}_{det}$ is our novel detector loss and $\frak{H}_1, \frak{H}_2$ are the height maps obtained by forwarding the pair of images through our backbone. 

At inference time, the keypoints are simply obtained by performing a fast non-maximum suppression algorithm that selects the locations corresponding to a local maximum of the height map with a value above a given threshold $\gamma$. Fig. \ref{fig:pipeline} summarizes the training and inference steps.

\section{Experiments}

\subsection{Dataset and Metrics}

We assessed the capability of our method to predict repeatable keypoints using the well-established HPatches benchmark \cite{balntas2017hpatches}. This dataset comprises 116 scenes, split into 696 images, with the first 57 scenes emphasizing variations in illumination and the subsequent 59 containing changes in viewpoint. Each sequence in the dataset comprises image pairs of increasing difficulty. We focus on this dataset, given that it represents a classical, longstanding benchmark for the task of keypoint detection, to assess the validity of our framework.

Regarding evaluation, our main concern is comparing the quality of the extracted keypoints for different detectors. Thus, as a metric, we use the formulation of repeatability proposed in \cite{VlbDet18}, which evaluates the consistency of keypoints across different images while overcoming the issues of previous definitions of repeatability, which can bias toward detecting clusters of keypoints \cite{rey2015comparing}. This adaptation is detailed in \ref{app:rev_rep} and aims to assess the unique association of keypoints by preventing a single keypoint from matching multiple counterparts, thus quantifying the proportion of keypoints that are each other's nearest neighbors in the corresponding images and are closer than a predefined distance threshold.

\subsection{Baselines}

In our study, we benchmark our model against a range of established detectors and state-of-the-art models to ensure a comprehensive evaluation:

\begin{itemize}[noitemsep,topsep=1pt]
\item \textbf{SIFT}: a handcrafted detector designed to be robust against scale changes. 
\item \textbf{D2-Net}: employs a multi-scale inference approach, detecting local maxima across multiple output maps. 
\item \textbf{R2D2}: an unsupervised detector that uses multi-scale inference. 
\item  \textbf{SuperPoint}: a semi-supervised detector trained to generalize to real images from labeled synthetic shapes. 
\item \textbf{DISK}: utilizes a probabilistic formulation to jointly model detection and matching. 
\item \textbf{ALIKED}: features deformable convolution, adapting receptor fields to the keypoints' support.
\end{itemize}

\begin{table}
\begin{center}
\begin{adjustbox}{width=.7\linewidth}
\begin{tabular}{lcccc}
\toprule

 Method & Avg & 75\% & 50\% & 25\%\\

\hline

D2-Net    & 24.6 & 31.9 & 19.2 & 22.8 \\
R2D2      & 48.5 & 55.7 & 56.2 & 33.7 \\
SIFT      & \textbf{63.6} & \underline{75.9} & \textbf{64.8} & \textbf{50.2} \\
SuperPoint& 60.6 & 73.3 & \underline{63.0} & \underline{45.6} \\
DISK      & 56.0 & 71.8 & 57.4 & 38.8 \\
ALIKED    & 18.7 & 24.2 & 16.5 & 15.4 \\
\textbf{MorseDet (ours)}  & \underline{62.2} & \textbf{82.2} & \underline{63.0} & 41.3 \\

\bottomrule
\end{tabular}
\end{adjustbox}
\caption{\label{tab:scales} Repeatability of the detector on resized HPatches images as the scale factor progressively reduces. The \textbf{best} and \underline{second-best} results are indicated in each column.}
\end{center}
\end{table}

\subsection{Implementation Details}
\label{details}

For training, we employed AdamW \cite{loshchilov2018decoupled} as an optimizer, 
with batches of 8 pairs of images with resolution 208 $\times$ 208.
To generate image pairs, we follow the protocol from R2D2, using the same training datasets (WASF).
Hyperparameter search and early stopping are performed on the validation split of MegaDepth \cite{li2018megadepth} used in \cite{detone2018superpoint}.
The final hyperparameters configuration included $\alpha=10$, weight decay equal to $0.005$, and repeatability threshold $\gamma = 0.7$ for inference.
The training process on a single TITAN X GPU with 12GB of VRAM concluded in approximately 10 hours until convergence.

\begin{figure}[!htb]
\begin{center}
    \minipage{0.17\textwidth}
      \includegraphics[width=\linewidth]{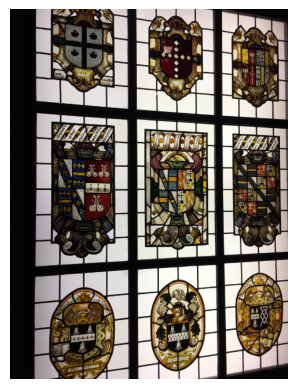}
      \caption*{\textbf{(a)}}
    \endminipage
    \hspace{0.5cm}
    \minipage{0.17\textwidth}
      \includegraphics[width=\linewidth]{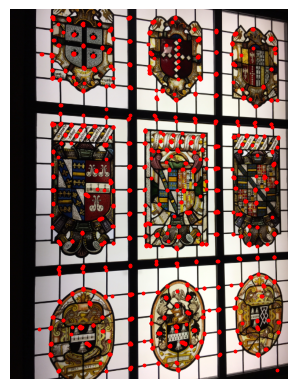}
      \caption*{\textbf{(b)}}
    \endminipage
\end{center}
\begin{center}
    \minipage{0.17\textwidth}%
      \includegraphics[width=\linewidth]{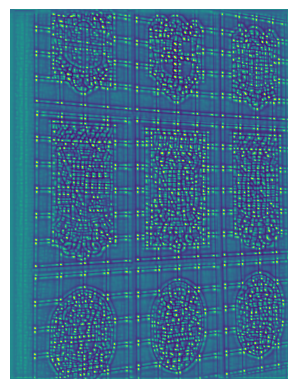}
      \caption*{\textbf{(c)}}
    \endminipage
    \hspace{0.5cm}
    \minipage{0.17\textwidth}%
      \includegraphics[width=\linewidth]{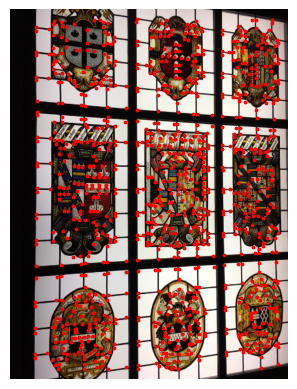}
      \caption*{\textbf{(d)}}
    \endminipage
\end{center}
\begin{center}
    \minipage{0.17\textwidth}%
      \includegraphics[width=\linewidth]{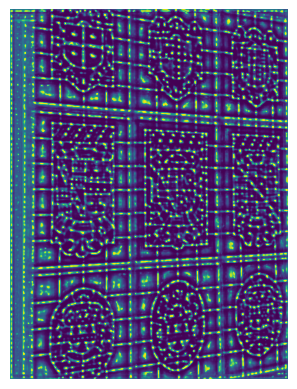}
      \caption*{\textbf{(e)}}
    \endminipage
    \hspace{0.5cm}
    \minipage{0.17\textwidth}%
      \includegraphics[width=\linewidth]{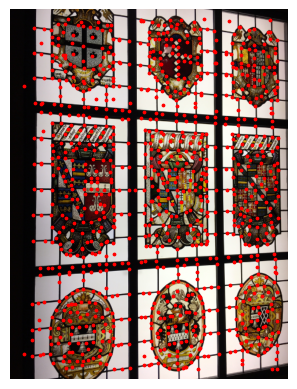}
      \caption*{\textbf{(f)}}
    \endminipage
\end{center}
\vspace{-0.2cm}
\caption{\label{fig:r2d2}(a) HPatches image. (b) R2D2 multi-scale inference keypoints. (c) MorseDet height map. (d) MorseDet keypoints. (e) R2D2 repeatability map. (f) R2D2 single-scale inference keypoints. R2D2 tends to detect keypoints at a fixed resolution, creating artifacts along edges and in untextured areas. In contrast, MorseDet better adapts its keypoints to the scale of the image content. While multi-scale inference can improve R2D2, it often results in redundant predictions.}
\vspace{-0.3cm}
\end{figure}

\begin{figure*}[!htb]

\minipage{0.19\textwidth}
  \includegraphics[width=\linewidth]{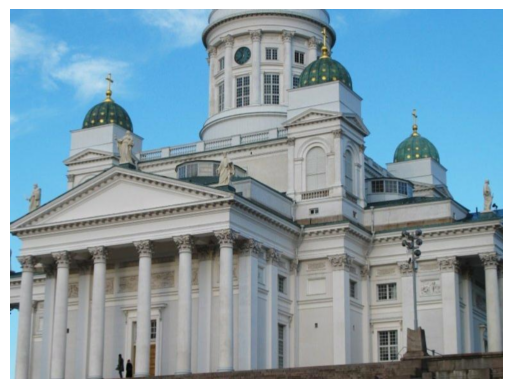}
  \caption*{\textbf{(a)}}
\endminipage
\hfill
\minipage{0.19\textwidth}
  \includegraphics[width=\linewidth]{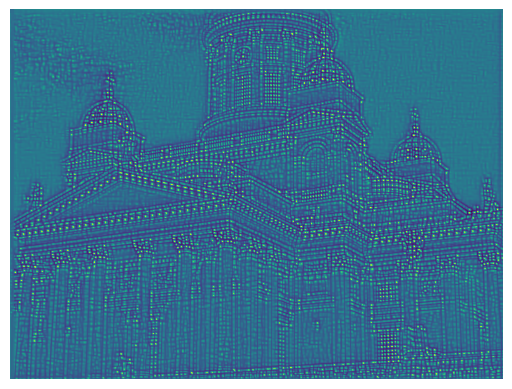}
  \caption*{\textbf{(b)}}
\endminipage
\hfill
\minipage{0.19\textwidth}%
  \includegraphics[width=\linewidth]{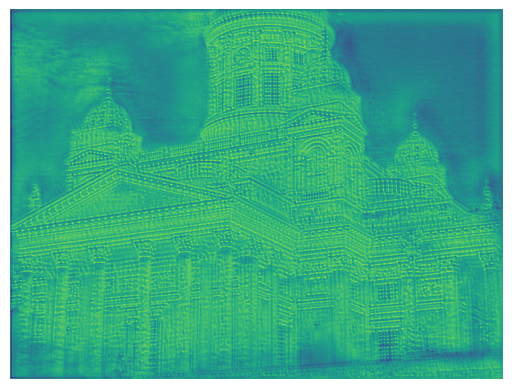}
  \caption*{\textbf{(c)}}
\endminipage
\hfill
\minipage{0.19\textwidth}%
  \includegraphics[width=\linewidth]{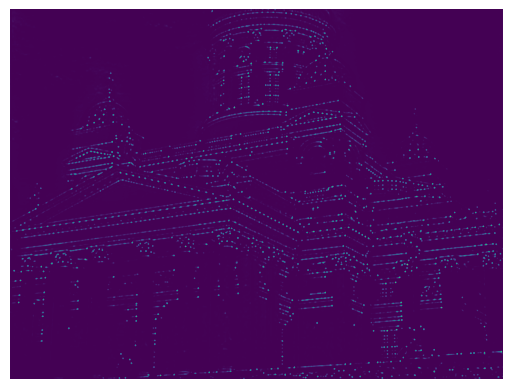}
  \caption*{\textbf{(d)}}
\endminipage
\hfill
\minipage{0.19\textwidth}%
  \includegraphics[width=\linewidth]{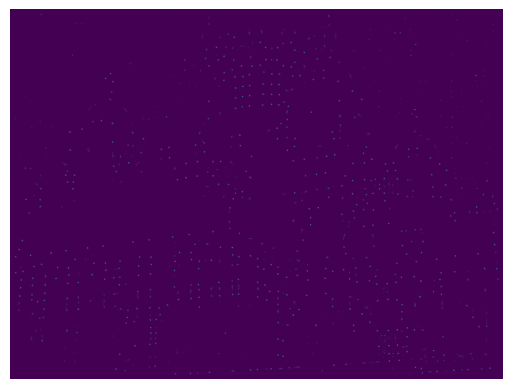}
  \caption*{\textbf{(e)}}
\endminipage

\hspace*{\fill}
\minipage{0.19\textwidth}%
  \includegraphics[width=\linewidth]{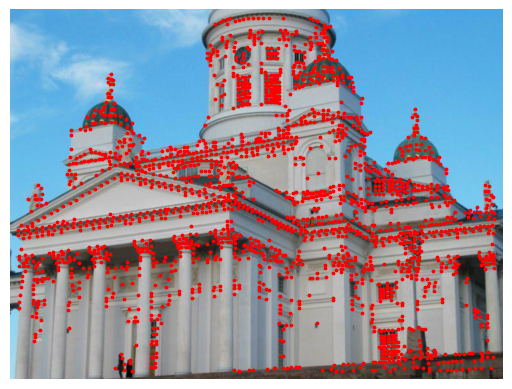}
  \caption*{\textbf{(f)}}
\endminipage
\hspace*{\fill}
\minipage{0.19\textwidth}%
  \includegraphics[width=\linewidth]{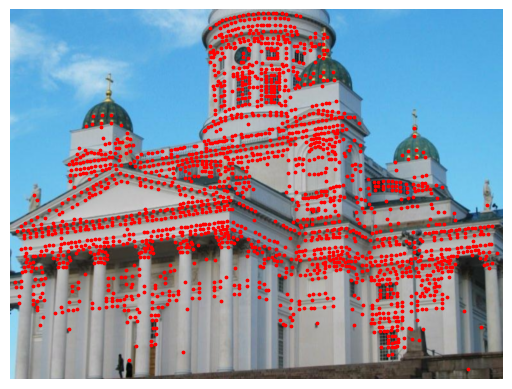}
  \caption*{\textbf{(g)}}
\endminipage
\hspace*{\fill}
\minipage{0.19\textwidth}%
  \includegraphics[width=\linewidth]{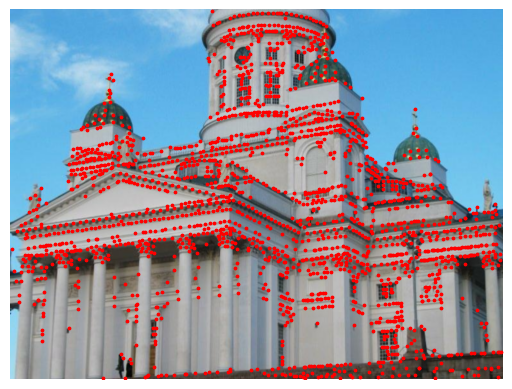}
  \caption*{\textbf{(h)}}
\endminipage
\hspace*{\fill}
\minipage{0.19\textwidth}%
  \includegraphics[width=\linewidth]{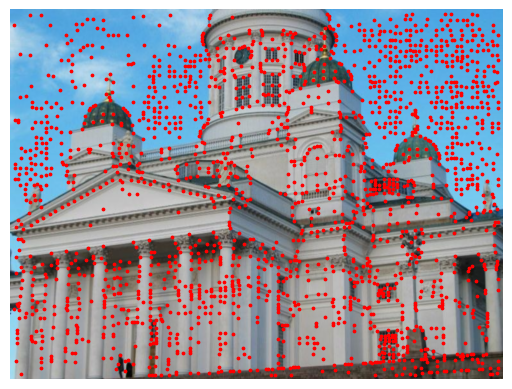}
  \caption*{\textbf{(i)}}
\endminipage
\hspace*{\fill}

\caption{\label{fig:mega}(a) Input image. (b) MorseDet, (c) DISK, (d) ALIKED, (e) SuperPoint scalar map. (f) MorseDet, (g) DISK, (h) ALIKED, (i) SuperPoint keypoints. DISK tends to generate equally-spaced keypoints, even in untextured areas. ALIKED predominantly produces keypoints at a fixed resolution. SuperPoint identifies the most clear features while overlooking others (to the extent of favoring the sky). MorseDet detects keypoints across various scales, and its height map reveals structural information from the input.}

\end{figure*}

\subsection{HPatches Benchmarks}
\label{sec:hp_bench}
\textbf{Detector Repeatability}

In this experiment, we evaluate the detector repeatability across changes in point of view and illumination on the common benchmark HPatches. Following \cite{revaud2019r2d2}, we provide results across different values for the maximum number of detected keypoints allowed. The results are shown in the tab. \ref{tab:hprepeat}, where the metrics are averaged across all thresholds up to 5px.

We can see that MorseDet's keypoints achieve consistently good performances, regardless of the number of keypoints or settings (\ie illumination and viewpoint changes), being either best or second best across the table.
Some other methods perform competitively with MorseDet under specific settings, although none is competitive in all cases.
Notably, DISK has strong results with a high number of keypoints, and SIFT is second best with fewer keypoints under viewpoint changes but performs poorly under illumination changes.
On average, SuperPoint is second-best.

\textbf{Scale Repeatability}
\label{sec:scale_rep}

We posit that models employing a fixed-size window approach for keypoint modeling during training learn to predict keypoints at a specific frequency. Building on this premise, such models may struggle to consistently replicate keypoints under rescaling transformations. To study this idea in isolation, we designed the following experiment using the images of HPatches. We evaluated for every method the repeatability metric between every image resized to 1000$\times$1000, and the image resized to smaller sizes to have approximately 75\%, 50\%, and 25\% the pixel area of the original image. As the number of keypoints deeply influences repeatability, we limit keypoints to 500, to ensure that every method uses the same number of keypoints at every scale for fair comparisons, thus also measuring how the methods can prioritize their most robust keypoints. The metrics are summarized in the tab. \ref{tab:scales} by their average above all the thresholds till 5px.

The results show that MorseDet obtains second-best results on average after SIFT.
In particular, MorseDet shines with 75\% image resize (\ie to images of 750$\times$750), outperforming the second best method, SIFT, by 6.3 points.
For extreme scale changes (\ie., 25\% of the original resolution), the best model is SIFT, which is a handcrafted detector built to be scale-invariant, followed by SuperPoint and MorseDet. 
Overall, the only learnable model competitive with MorseDet is SuperPoint, which benefits from a human-informed prior on keypoints. Notably, despite SIFT being proposed nearly two decades ago, it still outperforms modern detectors in this setup; MorseDet performs significantly better than every other learnable method in this task.
This is a direct consequence of the fact that previous learnable methods lack a principled framework for modeling local maxima, which is our method's core contribution.

\subsection{Qualitative Results}

In fig. \ref{fig:r2d2}, we compare the height map generated by MorseDet and the repeatability map produced by R2D2, using a HPatches image. The figure highlights that MorseDet exhibits superior adaptability to the context compared to the constraints posed by a patch-wise unsupervised detector loss. The repeatability map of R2D2 shows a bias towards detecting keypoints at a fixed resolution, potentially resulting in the exclusion of some features and the generation of artifacts, especially along edges and in untextured areas. In contrast, MorseDet can adjust its keypoints according to the image content, effectively detecting features at both large and fine-grained scales. Although R2D2 partially mitigates some of these issues through multi-scale inference, it tends to produce redundant predictions by clustering keypoints around the same semantic feature while potentially missing other points of interest.

In fig. \ref{fig:mega}, we compare the outcomes of MorseDet with other detectors based on scalar maps, namely SuperPoint, DISK, and ALIKED, using an image from the Megadepth dataset. The figure illustrates inherent limitations in all previous approaches. Specifically, DISK exhibits poor localization, densely detecting keypoints on buildings. SuperPoint demonstrates a limited understanding of keypoints, correctly identifying the most evident features (e.g. sharp corners in clear areas). Still, it favors noisy keypoints in untextured areas rather than finding additional keypoints in relevant structures. ALIKED performs better, but a hard-coded resolution bias persists in its scalar map. Conversely, our method does not suffer from these limitations. Moreover, it exhibits an attractive emergent property by effectively capturing much of the input structure in its height map.

\subsection{Ablation Study}

To better understand the design of the loss function and the respective effects of the boundary similarity and the persistence terms, we conducted an ablation study on the effect of training with different values of $\alpha$. The results regarding the repeatability of the HPatches viewpoint data split are reported in the tab. \ref{tab:alpha}. 

The results show that the model's performance is poor in the absence of the $Sim$ loss term (\ie $\alpha=0$). Conversely, increasing $\alpha$ increases performance across all considered keypoint quantities. At $\alpha=20$ the repeatability continues to rise when the number of keypoints is limited but does not gain additional benefits from allowing a more significant number of keypoints,  suggesting a trade-off imposed by the parameter's value, balancing the quantity and quality of detected keypoints. 

\begin{figure}[!htb]
\begin{center}
    \minipage{0.17\textwidth}
      \includegraphics[width=\linewidth]{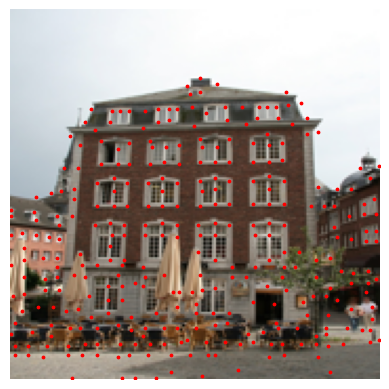}
      \caption*{\textbf{(a)}}
    \endminipage
    \hspace{0.5cm}
    \minipage{0.17\textwidth}
      \includegraphics[width=\linewidth]{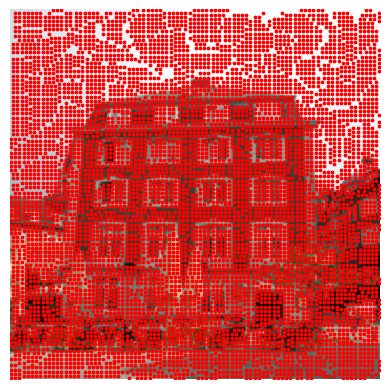}
      \caption*{\textbf{(b)}}
    \endminipage
\end{center}

\vspace{-0.2cm}
\caption{\label{fig:alpha} The images display all detected keypoints for a $208\times208$ training image, comparing
outcomes from (a) the actual model with  $\alpha=10$ and (b) a model trained with $\alpha=0$.}
\vspace{-0.3cm}
\end{figure}

Indeed, without the $Sim$ term (\ie $\alpha=0$), the loss simplifies to the maximization of squared persistence, thus amounting to seeking an output feature map with as many local maxima as possible, leading to a trivial and uninformative solution, a grid of 1s surrounded by 0s in every $3\times3$ image patch, disregarding the input image value. The limitation of the model and training process impedes achieving an ideal grid pattern exactly, as shown in fig. \ref{fig:alpha}. Nonetheless, training without the $Sim$ component yields an output that resembles the trivial solution. On the contrary, setting a sensible value for $\alpha$ enables learning repeatability from the data.

\begin{table}
\begin{center}
\begin{adjustbox}{width=.7\linewidth}
\begin{tabular}{lcccccc}
\toprule

$\alpha$ & & 250 & 500 & 1000 & 2000 & 4000 \\

\hline

0 & & 1.0 &	1.6 & 1.9 &	3.0 & 5.9 \\
5 & & 34.8 & 37.5 & 40.3 & 42.9 & \underline{45.1} \\
10 & & \underline{40.6} & \underline{42.8} & \underline{44.6} & \textbf{46.1} & \textbf{47.2} \\
20 & & \textbf{44.0} & \textbf{44.8} & \textbf{45.3} & \underline{45.1} & 44.9 \\

\bottomrule
\end{tabular}
\end{adjustbox}
\caption{\label{tab:alpha} Repeatability on HPatches viewpoint across various value for the maximum
number of keypoints allowed, for models trained with different values of $\alpha$. The \textbf{best} and \underline{second-best} results are indicated in each column.}
\end{center}
\end{table}

\section{Conclusion and Limitations}

In this study, we introduced the first training pipeline driven entirely by a loss function based on persistent homology. We based our approach on the observation that Morse theory and persistent homology are naturally suited for modeling the local extrema of a scalar map in a scale-independent way. Through the discrete Morse theory framework, we showed that these concepts can be applied to model the salient points of a digital image in a differentiable manner suitable for keypoint detection. Our experiments on the classical HPatches benchmark demonstrated promising results that validate our methods. We believe this work can serve as a stepping stone for future research in extending the applications of topological data analysis to the fields of computer vision and image matching.

However, this novelty also presents certain limitations. Due to the limited popularity of discrete Morse theory within the deep learning literature, more open-source software must be needed for efficient implementation. Our current implementation of the loss function, aligned with the methodologies of \cite{robins2011theory}, operates on a CPU. Despite this, as detailed in sec. \ref{details}, training takes roughly 10 hours.

\section*{Acknowledgements}

This study was carried out within the project FAIR - Future Artificial Intelligence Research - and received funding from the European Union Next-GenerationEU (PIANO NAZIONALE DI RIPRESA E RESILIENZA (PNRR) – MISSIONE 4 COMPONENTE 2, INVESTIMENTO 1.3 – D.D. 1555 11/10/2022, PE00000013). This manuscript reflects only the authors’ views and opinions; neither the European Union nor the European Commission can be considered responsible for them.

\section*{Impact Statement}

This paper presents MorseDet, a work that aims to advance the field of Machine Learning. Our work has many potential societal consequences, none of which we feel must be specifically highlighted here.

\bibliography{camera_ready}
\bibliographystyle{icml2024}


\clearpage
\appendix

\section{Algebraic Topology}

\label{sec:algtop}

\subsection{Cubical Complex}
\label{sec:cubical}

A cubical complex is a \emph{finite} family $\mathcal{K}$ of objects $Q\subset \mathbb{R}^d$, such that: for all $Q\in\mathcal{K},$ there is a set of integers $I_Q:=\{l_1,\dots,l_d\}$ such that $Q=I_1\times\dots\times I_d,$ with $I_j=[l_j,l_j+1]$ or $I_j=[l_j,l_j]$; and, if $P,Q\in\mathcal{K}$ then either $P\cap Q=\emptyset$ or $P\cap Q\in\mathcal{K}.$
When $I=[l,l],$ it is called \emph{degenerate}, and the number of non-degenerate intervals in $Q$ is its dimension, while $d$ is usually called its embedding number. The $0-$dimensional cubes are points, $1-$dimensional cubes are edges, $2-$dimensional cubes are squares, and so on. We denote by $\mathcal{H}_k$ the subset of $k-$dimensional cube in $\mathcal{K},$ thus, for example $\mathcal{K}_0$ are the $0-$cubes, or vertices, or points.
Let $P\subset Q\in\mathcal{K},$ then $P$ is called a \emph{face} of $Q,$ and, if the inclusion is proper, $\dim{P}\leq \dim{Q}-1.$
A cubical complex $\mathcal{K}$ is a partially ordered set (poset) via inclusion and, if $(\mathcal{P}, \leq_{\mathcal{P}})$ is another poset with $f:\mathcal{K}\to \mathcal{P}$ a monotonically not decreasing map, that is $P\subset Q$ implies $f(P)\leq_{\mathcal{P}} f(Q),$ then it is possible to create a sublevel filtration of $\mathcal{K}$ as follows: $\emptyset \subset \mathcal{K}_1\subset\mathcal{K}_2\dots\subset\mathcal{K}_t=\mathcal{K} $
where $\mathcal{K}_s:=\{Q\in \mathcal{K}\,:\, f(Q)\leq_{\mathcal{P}} p_s\in\mathcal{P}\}.$
Topological invariants of these sublevel sets and their behavior along the filtration are some of the main topics in topological data analysis. In particular, they have been studied in Morse theory, discrete Morse theory, and persistent homology.

It is customary to represent $2D-$greyscale digital images as cubical complexes as follows: $0-$cubes are the images pixels laying on the vertices of an integral rectangular lattice in $\mathbb{R}^2;$ $1-$dimensional cubes are the edges connecting pixel that differ by $1$ in precisely one coordinate; $2-$dimensional cubes, $i.e.$ squares, are the obvious ones. 
Let $I:\{0-cubes\}\to [0,1]$ be the function assigning to each pixel its values. Then, we can associate $I$ with a function $f:\mathcal{K}_I\to [0,1]$ by $f(Q)=max_{P\in\mathcal{K}_0\,:\, P\subseteq Q}I(P).$
The complex $\mathcal{K}_I,$ with $f_I$ and the corresponding filtration will be called the \emph{cubical complex associated to I} or, more simply, the \emph{I-complex}.
We will see how to take advantage of this representation in a moment, but we need some further mathematical devices before proceeding. 
\subsection{Homology and Persistence}
Homology is a topological invariant whose story goes back to the seminal work of H.Poincar\`{e} \emph{Analysis Situs} published in $1895$. For our setting, discussing homology for dimensions equal to zero and one only would suffice. In these cases, homology will count the number of connected components of a complex (dimension zero) and the number of closed one-dimensional paths that cannot be shortened (cycles or one-dimensional hole in the complex. 
Homology is defined via two objects: first, for $k=0,1,2$, we consider the vector spaces $C_k(\mathcal{K})$ with bases in bijection with $\mathcal{K}_k$, notice that there are some subtleties regarding the orientation of the cube here, see \cite{edelsbrunner2022computational}. Then, we introduce the boundary maps $\partial_k: C_k\to C_{k-1}$, which is the algebraic representation of the operation of taking the boundary of a cube. For example $\partial_1([l,l+1])=[l+1]-[l]$ or $\partial_2([l,l+1]\times[l,l+1])=[l,l+1]\times[l]+[l+1]\times[l,l+1]-[l,l+1]\times [l+1] - [l]\times[l,l+1].$
It is not difficult to check that $\partial_k\partial_{k+1}=0$ for all suitable $k$, and this implies that the image of $\partial_k$, called $B_k$ for \emph{boundaries}, is a subspace of the kernel of $\partial_{k-1}$, denoted $Z_k$ for cycles (zyklen). Their quotient $H_k:=Z_k/B_k$ is called the $k-$th homology group (of $\mathcal{K}$). 
When one has a map of complexes $\mathcal{K}\to \mathcal{H}$, it induces a linear mapping $H_k(\mathcal{K})\to H_k(\mathcal{H})$ for all $k$ and one can check what happens to connected component and cycle of $\mathcal{K}$ when mapped to $\mathcal{H}.$
Of paramount importance in our setting is that the sublevel filtration of the $I-complex$ induces a sequence of linear mappings on the homology groups. The space $\mathbb{H}_k(\mathcal{K}_I, f_I)=\bigoplus_{f_I([a,b])\,:\, [a,b]\in \mathcal{K}_I}H_1(\mathcal{K}_I(f[a,b]))$is called the $k-$th \emph{persistent homology group} of the filtered space $(\mathcal{K}_I, f_I)$. It is generated by persistent cycles as a module over the ring of polynomial in one variable, and, in the case of the cubical complex associated with an image $I$, it holds that the generators of its first persistent homology group are in one to one correspondence with the local maxima of the image, see \cite{edelsbrunner2022computational}.

\section{Differentiability}
\label{sec:diff}

From the more general results of \cite{leygonie2021framework, carriere2021optimizing}, given a cubical complex $K$, there is a stratified function $F$,  denoting with $\Omega(\mathbb{R}^{HW})$ the set of its strata, such that:

\begin{itemize}[noitemsep,topsep=1pt]
    \item on each stratum $S \in \Omega(\mathbb{R}^{HW})$ the function restricted to $S$ behaves as a permutation $P_S$.
    \item if $\frak{H}  \in \mathbb{R}^{HW}$ has all distinct values, there is a stratum $S$ such that $\frak{H} \in S$,  where $P_S$ returns the $m$ pairs of critical times associated with its local maxima in increasing order of birth time, followed by all the other $n$ unpaired values in increasing order $F(\frak{H})=(b_1, d_1, ..., b_{m}, d_{m}, u_1, ..., u_{n})$.
\end{itemize}

Intuitively, if $\frak{H}$ has all distinct values and $d$ is the minimum distance between them, in a neighborhood constituted by the open
ball  $\mathcal{B}(\frak{H}, d)$ the location of its critical times cannot change. Consequently, our loss's persistence and boundary similarity terms can be expressed through composition with a fixed permutation $P_S$ on a neighborhood of $\frak{H}$, making them differentiable almost everywhere. Suppose multiple $\frak{H}$ entries have the same value. In that case, we can remove the problem by considering a perturbation that makes all the values distinct while maintaining any other order relation. For instance, $\frak{H}_\epsilon = \frak{H} + \epsilon V$ where $V[i, j] = \frac{i + Ij}{2IJ}$ and $\epsilon$ is smaller than $d$ the minimum positive difference between different entries. For $\epsilon \in (0, d)$, all $H_\epsilon$ belong to the same stratum that corresponds to a fixed permutation $P_S$, allowing us to compute a directional derivative by arbitrarily choosing a trajectory $V$. In practice, we can arbitrarily break ties if needed.

\section{Revisited Repeatability}
\label{app:rev_rep}

Following the protocol established by  \cite{mikolajczyk2005performance}, repeatability is a key metric in literature for assessing the detection of reproducible keypoints. We adopt the version from \cite{detone2018superpoint}, which aligns with current point-based prediction methods. For clarity, we detail this metric: given two sets of predicted keypoints $A, B$ from a pair of images $I_1, I_2$ related by a homography $U$, a keypoint $x \in A$ is positively referenced in $B$ if $\min_{y\in B}||x-U^{-1}(y)||$ is less than a threshold $\epsilon$ , where $U^{-1}(y)$ is the projection of $y$ through the ground truth homography. The repeatability score is the average number of keypoints with a positive reference, typically assessed within covisible areas, acknowledging the detectors' limitation of not knowing a priori which regions will be matched.

However, repeatability is influenced by the number of extracted keypoints. For instance, a uniformly distributed grid of keypoints can artificially inflate the score. To mitigate this, we varied the maximum number of keypoints in our experiments as in \cite{revaud2019r2d2}. Despite this, the metric could still favor detectors producing clustered keypoints, as noted by \cite{rey2015comparing, VlbDet18}. The work in \cite{VlbDet18} suggested a method that allows keypoints to match at most one time: computing repeatability based on matches from an optimally constructed bipartite graph, minimizing the sum of a cost function based on distance, with a proposed greedy approximation for the optimization problem.

Our revisited repeatability is a simplification that further requires keypoints to be mutually nearest neighbors, \ie 
\begin{equation}
x=\argmin_{x'\in A}||y-U(x')||
\end{equation}
and
\begin{equation}
y=\argmin_{y'\in B}||x-U^{-1}(y')||
\end{equation}

This addition helps unequivocally associate underlying features and penalizes redundant detections, favoring detectors more prone to cover image features with a limited number of keypoints comprehensively.

\end{document}